# A New Approach for Finding the Global Optimal Point Using Subdividing Labeling Method (SLM)

**MasoumehVali**

Department of Mathematics, Dolatabad Branch, Islamic Azad University, Isfahan, Iran
E-mail: vali.masoumeh@gmail.com

**Abstract**

In most global optimization problems, finding global optimal point inthe multi-dimensional and great search space needs high computations. In this paper, we present a new approach to find global optimal point with the low computation and few steps using subdividing labeling method (SLM) which can also be used in the multi-dimensional and great search space. In this approach, in each step, crossing points will be labeled and complete label polytope search space of selected polytope will be subdivided after being selected. SLM algorithm finds the global point until h (subdivision function) turns into zero. SLM will be implemented on five applications and compared with the latest techniques such as random search, random search-walk and simulated annealing method. The results of the proposed method demonstrate that our new approach is faster and more reliable and presents an optimal time complexity O ($\log_2^n$).

## 1 Introduction

Optimization problem is a famous field of the science, engineering and technology. In the majority of these problems, it is necessary to compute the global optimum (or a good approximation) of a multivariable function. The variables which define the optimized function can be continuous or discrete and; additionally, they often have to satisfy certain constraints. The global optimization problems belong to the complexity class of NP-hard problems. Such problems are very difficult to solve and the traditional descent optimization algorithms based on local information are inadequate to solve them. Global optimization has attracted a lot of attention in the last ten years, due to the success of new algorithms for solving large classes of problems from diverse areas such as computational chemistry and biology, structural optimization, computer sciences, operations research, economics and engineering design and control.

SLM doesn't need to handle derivatives to search a continuous domain like other methods such as RS, RSW and SA. Random Search (SR) is the Global Optimization and direct search method in Stochastic Optimization field. [1, 2]

## 2. Related Work

In 1953, Metropolis et al. [3] developed a Monte Carlo method for "calculating the properties of any substance which may be considered as composed of interacting individual molecules". With this so-called "Metropolis" procedure stemming from statistical mechanics, the manner in which molten metal crystallizes and reaches equilibrium in the process of annealing can, for instance, be simulated. This inspired Kirkpatrick et al. [4] to develop the Simulated Annealing (SA) global optimization algorithm in the early 1980s and to apply it to various combinatorial optimization problems. The effectiveness of a hybrid of simulated annealing (SA) and the simplex algorithm for optimization of a chromatographic separation has been examined. The method has been used to solve optimization problems for separations in batch and continuous chromatographic systems under isocratic and gradient conditions. [5]

In this paper, the following optimization problems such as f $(x_1, x_2, ..., x_m)$ where each $x_i$ is a real parameter object to $a_i \leq x_i \leq b_i$ for some constants $a_i$ and $b_i$ have been discussed. We present a new method based on SLM. This approach can earn globaloptimal points (max/min) in the few steps while it does not require derivation. The proposed method uses the multi-dimensional and great search space. In Section2, we present related works which have been done yet. The description of SLM schema and comparison of this proposed method with other methods such as RS, RSW and SA will be discussed in Section 3. In Section 4, the time complexity of SLM has been computed. Finally, we have presented the conclusion of the work in section 6.

## 3. Problem Definition for the SLM

In this section, we define the model of SLM and implement several test problems such as the second and fifth De Jong's functions by the SLMAlgorithm. At last, the results of these implementations will be compared with those of other methods (RS, RSW, SA).

### 3.1 SLM Algorithm

Consider the functionf( $x_1, x_2, x_3 ....., x_n$) with constraint $a_i \leq x_i \leq b_i$. We want to earn the global optimal point based on the following serial algorithm SLM:

**Step1:** Draw the diagrams for $x_i = b_i \text{ and } x_i = a_i$ for i=1, 2,..., n ; then, we find crossing points which equal $n^2$ and h= min$\frac{|x_i|+|x_j|}{2}$ for $1 \leq i, j \leq n$.

**Step2:** Suppose that the point$(a_1, b_1, b_2 \ldots b_{n-1})$ is one of the crossing points. Consider the values of $\pm h$ gained by step1 and then do the algebra operations on this crossing point as shown in Table 1. Their maximum number in dimensional space is $2 * (\binom{n}{1} + \binom{n}{2} + \binom{n}{3} + \cdots + \binom{n}{n}))$ .

Table 1: The number of points produced by algebra operation on a primary point.

| |
|---|
| $(a_1 \pm h, b_1, b_2 \ldots b_{n-1})$ |
| $(a_1, b_1 \pm h, b_2 \ldots b_{n-1})$ |
| $\vdots$ |
| $(a_1, b_1, b_2 \ldots b_{n-1} \pm h)$ |
| $\vdots$ |
| $(a_1 \pm h, b_1 \pm h, b_2 \ldots b_{n-1})$ |
| $\vdots$ |
| $(a_1 \pm h, b_1 \pm h, b_2 \pm h, \ldots, b_{n-1} \pm h)$ |

**Step 3:** The function value is calculated for all points of step 2 and the value of these functions is compared with f$(a_1, b_1, b_2 \ldots b_{n-1})$ . At the end, we will select the point which has the minimum value and we will call it $(c_1, c_2, c_3 \ldots c_n)$.

**Note:** If we want to find the optimal global max, we should select the maximum value.

**Step 4:** In this step, the equation1 is calculated:

$$(c_1, c_2, c_3 \ldots c_n) - (a_1, b_1, b_2 \ldots b_{n-1}) = (d_1, d_2, d_3 \ldots d_n) \quad (1)$$

**Step 5:** According to the result of step 4, the point $(a_1, b_1, b_2 \ldots b_{n-1})$ is labeled according to equation2.

$$l \begin{cases} 0 & d_1 \geq 0, \ldots, d_n \geq 0 \\ 1 & d_1 < 0, d_2 \geq 0, \ldots, d_n \geq 0 \\ 2 & d_2 < 0, d_3 \geq 0, \ldots, d_n \geq 0 \\ \vdots & \\ N & d_n < 0 \end{cases} \quad (2)$$

**Step 6:** Go to step 7 if all the crossing points were labeled, otherwise repeat the steps 2 to 5.

**Step 7:** In this step, the complete labeling polytope is focused. In fact, a polytope will be chosen that has complete labeling in different dimensions as shown in Table 2.

Table 2: complete labeling in different dimensions

| Dimension | Complete Label |
|---|---|
| 2 | 0,1,2 |
| 3 | 0,1,2,3 |
|  |  |
| n | 0,1,2,3,…,n |

**Step 8:** In this step, all sides of the selected polytope (from step 7) are divided into 2 according to equation 3 and we repeat steps 3 to 7 for new crossing points.

$$h = \min\left\{\frac{|x_i|+|x_j|}{2}\right\} \quad 1 \leq i,j \leq n \quad (3)$$

**Step 9:** Steps 2 to 8 are repeated to the extent that $h \to 0$ and the result is global min or max point.

## 3.2 Experimental Results:

In this part, we have five test problems implemented by SLM and at last, the results of algorithms: SLM, random search (RS), random selection–walk (RSW) and simulated annealing (SA) are compared in Tables 1 to 5.

### 3.2.1 Test Problem 1

Equation 4 is a continuous optimization problem.
$$minf(x_1, x_2) = x_1^2 + (x_2 - 0.4)^2 \tag{4}$$

$$-2 < x_i < 2, i = 1,2$$

The function achieves the minimum when $x_1 = 0$, and $x_2 = 0.4$. In this example, $h_i \in \{4, 2, 1, 0.5\}$ mutation probability $p_m = 1$. The completely label square obtains through the iteration, the search scope for both $x_1$, $x_2$ are (-2,2), (0,2), and finally (0,1) respectively (as shown in Figure 1 to Figure 3 and in table 3 to 5 ). During repetitions, squares are gradually contracting to (0, 0.5), if we started from $h_1 = 1$, we got closer answer i.e. (0, 0.4).

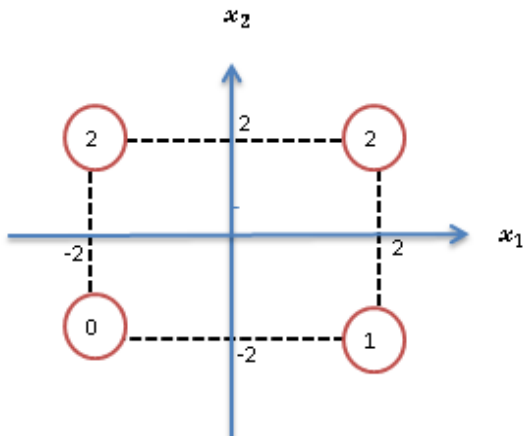

Figure 1: Initial Population of $f_1$

Table 3: Initial Population of $f_1$

| $h_1=4$, P(0): | $h_2=2$ | P(1): | $l(x)$ | Solution |
|---|---|---|---|---|
| (-2,2) |  | (0,0) | 2 |  |
| (2,2) |  | (0,0) | 2 |  |
| (-2,-2) |  | (0,0) | 0 | (0,0) |
| (2,-2) |  | (0,0) | 1 |  |

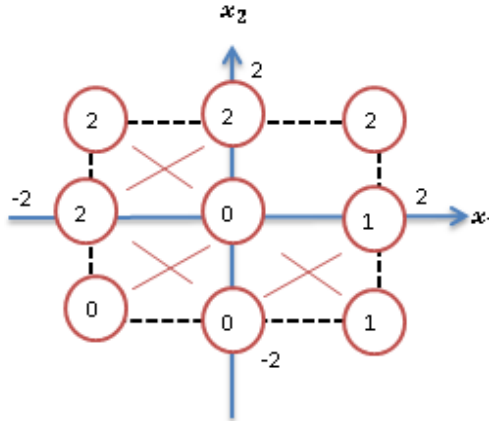

Figure 2: First generation of $f_1$

Table 4: First generation of $f_1$

| $h_2=2$ P(1): | $h_3=1$ | P(2): | $l(x)$ | Solution |
|---|---|---|---|---|
| (-2,2) | | (-1,1) | 2 | |
| (2,2) | | (1,1) | 2 | |
| (-2,-2) | | (-1,-1) | 0 | |
| (2,-2) | $\xrightarrow{M}$ | (1,-1) | 1 | (0,0) |
| (2,0) | | (1,0) | 1 | |
| (0,2) | | (0,1) | 2 | |
| (-2,0) | | (-1,0) | 0 | |
| (0,-2) | | (0,-1) | 0 | |

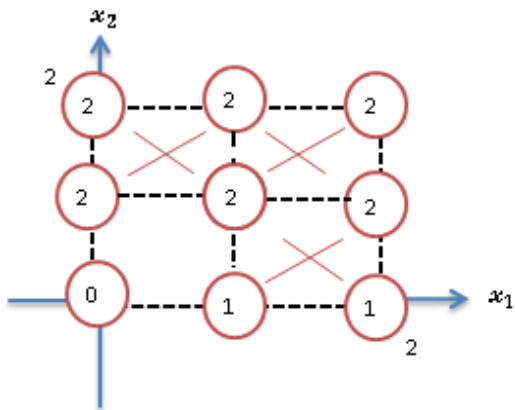

Figure 3: Second generation of $f_1$

Table 5: Second generation of $f_1$

| $h_3=1$ P(2): | $h_4=0.5$ | P(3): | $l(x)$ | Fixed point |
|---|---|---|---|---|
| (-1,1) | | (-0.5,0.5) | 2 | |
| (1,1) | | (0.5,0.5) | 2 | |
| (-1,-1) | | (-0.5,-0.5) | 0 | |
| (1,-1) | | (0.5,-0.5) | 1 | |
| (1,0) | | (0.5,0.5) | 1 | |
| (0,1) | $\xrightarrow{M}$ | (0,0.5) | 2 | (0,0.5) |
| (-1,0) | | (-0.5,0.5) | 0 | |
| (0,-1) | | (0,-0.5) | 0 | |
| (0,0) | | (0,0.5) | 0 | |
| (-1,2) | | (-0.5,1.5) | 2 | |
| (2,2) | | (1.5,1.5) | 2 | |
| (-2,1) | | (-1.5,0.5) | 2 | |

We compared SLM of problem 1 with other methods and their results are shown in table 7. Results show SLM earns global optimal the same as other methods but it bears lower steps and more precision than other methods.

Table 6: comparison between test problem #1 and other three

| Algorithms | Iteration | Optimal point | Best Point | Standard deviation |
|---|---|---|---|---|
| SLM | 6 | (0,0.4375) | | (0, 0.0375) |
| RS | 1000 | (0,0) | | (0,0.4) |
| $RSW(x^{initial} = (14.0356, 14.0356))$ | 500 | ( 0, 0. 49999996 ) | (0,0.4) | (0,0.09999996) |
| SA | 150 | (0,0.42) | | (0,0.02) |

### 3.2.2 Test problem2

Equation 5 is a nonlinear optimization problem with two continuous variables.

$$min\ f(x_1, x_2) = \cos\frac{\pi}{2}x_1 - \sin\frac{\pi}{2}x_2 \tag{5}$$

$$-7 < x_i < 7, \quad i = 1,2$$

This multimodal function has several local optimal points in its domain. The SLM earns each local and global optimal found in completely labeled squares. In this example these points have been gained, for $h_i \in \{6, 3, 1.5, 0.75\}$ while mutation probability $p_m = 1$, as shown in figure 6. Three first following generations have been shown in the first quarter of the coordinate system (see Figure 4 to 6).

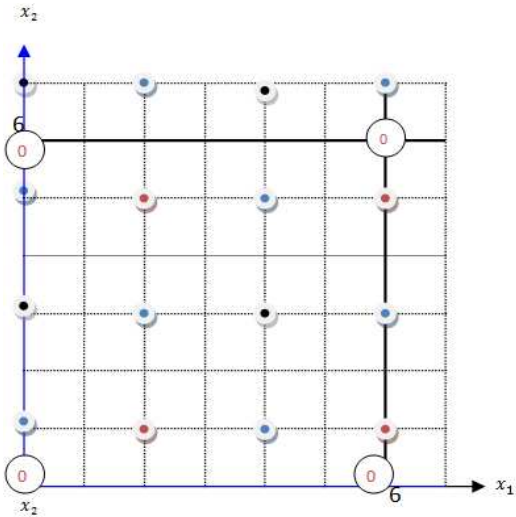

Figure 4: First generation of $f_3$

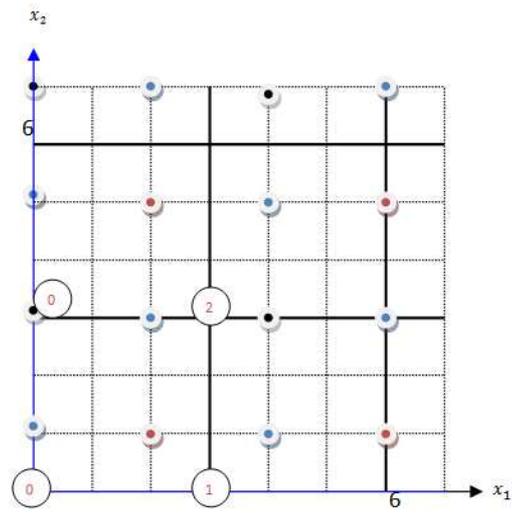

Figure 5: Second generation of $f_3$

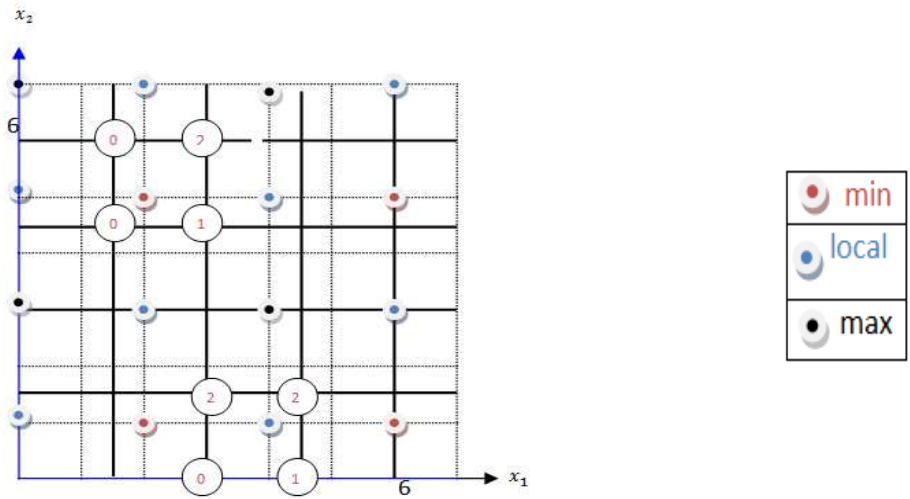

Figure 6: Third generation of $f_3$

Table 7: comparison between test problem #2 and other three methods

| Algorithms | Iteration | Optimal point | Best Point | Standard deviation |
|---|---|---|---|---|
| **SLM** | 7 | $(\pm 4k_1 \pm 1.78125, 2k_2+0.78125)$, $k_1=0,1, k_2=0,2$ | $(\pm 4k_1 \pm 2, 2k_2+1)$, $k_1=0,1, k_2=0,2$ | (0.21875, 0.21875) |
| **RS** | 1000 | $(\pm 4k_1 \pm 2.48, 2k_2+1.25)$, $k_1=0,1, k_2=0,2$ | | (0.48, 0.25) |
| **RSW**($x^{initial} =$ (18.048, 14.899)) | 500 | $(\pm 4k_1 \pm 1.999, 2k_2+.0999)$, $k_1=0,1, k_2=0,2$ | | (0.001, 0.001) |
| **SA** | 150 | $(\pm 4k_1 \pm 1.9499, 2k_2+.0999)$ $k_1=0,1, k_2=0,2$ | | (0.0501, 0.001) |

We compared the SLM of test problem 2 with other methods and their results are shown in table 7. Results show that the SLM earns the global optimal the same as other methods, but with lower steps and more precision than the other methods.

### 3.2.3 Test Problem3

Eequation 6 is a continuous optimization problem. In this test problem, SLM achieved the global optimal point in first step.

$$max\ f(x_1, x_2) = x_1^2 + (x_2 - 0.4)^2 \qquad (6)$$

$$-2 \leq x_i < 2, i = 1,2$$

The function achieves the maximum when $x_1 = -2$, and $x_2 = -2$.

Table 8: Initial Population of $f_3$

| $h_1 = 4$ P(0): | $h_2 = 2$ | P(1): | $l(x)$ | Solution |
|---|---|---|---|---|
| (2,2) | | (0,0) | 2 | |
| (-2,2) | $\xrightarrow{M}$ | (-2,0) | 2 | (-2,-2) |
| (-2,-2) | | (-2,-2) | 0 | |
| (2,-2) | | (0,-2) | 1 | |

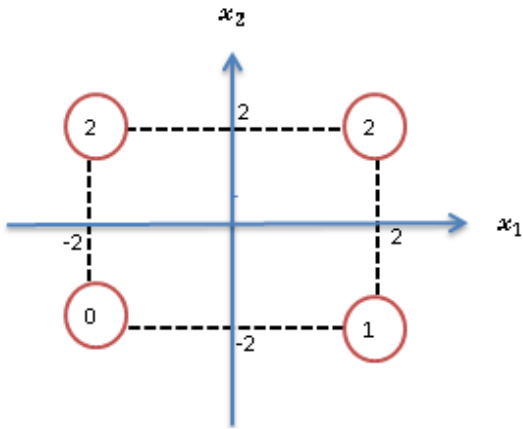

Figure 7: Initial Population of $f_3$

Table 9: Second generation of $f_4$

| $h_2 = 2$ P(1): | $h_3 = 1$ | P(2): | $l(x)$ | Solution |
|---|---|---|---|---|
| (2,0) | | (0,0) | 1 | |
| (-2,0) | | (-2,-1) | 2 | |
| (0,-2) | $\xrightarrow{M}$ | (-1,-2) | 1 | (-2,-2) |
| (0, 2) | | (-1,-2) | 2 | |
| (0,0) | | (-1,-1) | 2 | |

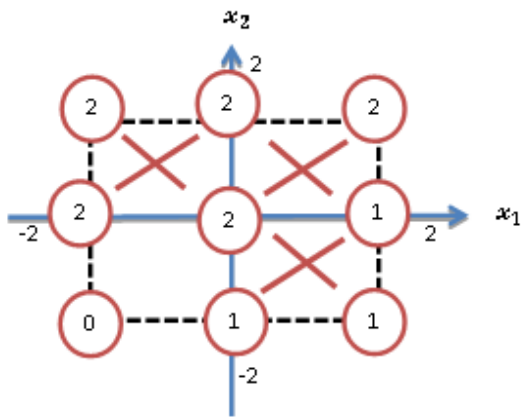

Figure 8: Second generation of $f_4$

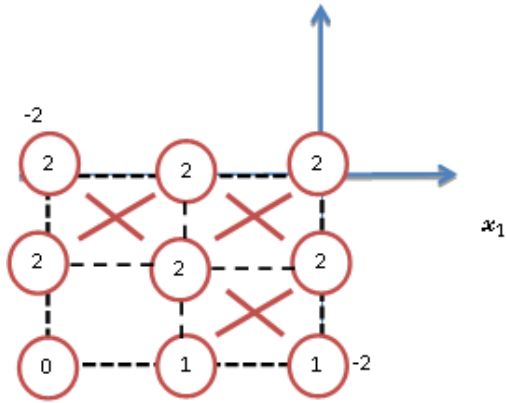

Figure 9: Third generation of $f_4$

Table 10: Third generation of $f_4$

| $h_3 = 1$ P(2): | $h_4 = 0.5$ | P(3): | $l(x)$ | Solution |
|---|---|---|---|---|
| (-1,0) | | (-1.5,-0.5) | 2 | |
| (-2,-1) | | (-2,-1.5) | 2 | |
| (0,-1) | $\xrightarrow{M}$ | (-0.5,-1.5) | 2 | (-2,-2) |
| (-1, -2) | | (-1.5,-2) | 1 | |
| (-1, -1) | | (-1.5,-1.5) | 2 | |

Table 11: comparison between test problem #3 and other three methods

| Algorithms | Iteration | Optimal point | Best point | Standard deviation |
|---|---|---|---|---|
| SLM | 10 | (-1.99609375,-1.99609375) | | (- 0.00390625, - 0.00390625) |
| RS | 1000 | (-1.5,-1.5) | | (-0.5,-0.5) |
| RSW($x^{initial} = (-14.0356, -14.0356)$) | 500 | (-1.987, -1.987) | (-2,-2) | (-0.013,-0.013) |
| SA | 150 | (-1.999099,-1.999099) | | (-0.000901,-0.000901) |

We compared the SLM of test problem 3 with other methods and their results are shown in table 11. Results show that the SLM earns global optimal the same as other methods, but with lower steps and more precision than the other methods.

### 3.2.4 Test problem 4

Problem (7) is the second De Jong function (Rosenbrock's saddle). [6]

$$f_2(x_0, x_1) = 100.(x_0^2 - x_1) + (1 - x_0)^2; \quad x_j \in [-5.12, 5.12] \quad (7)$$

Although $f_2(x_0, x_1)$ has just two parameters, it has the reputation of being a difficult minimization problem.
The minimum is $f_2(1,1) = 0$ that figures from 10 to 13 and tables from 12 to 15 are shown.

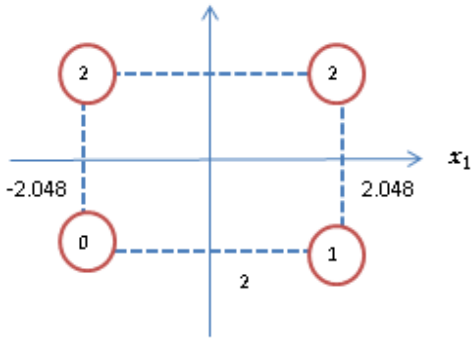

Figure 10: Initial Population of $f_2$

Table 12: Initial Population of $f_4$

| $h_1 = 4.096$ P(0): | $h_2 = 2.048$ | P(1): | $l(x)$ | Solution |
|---|---|---|---|---|
| (2.048,2.048) | $\xrightarrow{M}$ | (0,0) | 2 | (0,0) |
| (2.048,-2.048) | | (0,0) | 1 | |
| (-2.048,-2.048) | | (0,0) | 0 | |
| (-2.048,2.048) | | (0,0) | 2 | |

Table 13: The First generation of $f_2$

| $h_2 = 2.048$ P(1): | $h_3 = 1.024$ | P(2): | $l(x)$ | Solution |
|---|---|---|---|---|
| (0, 2.048) | $\xrightarrow{M}$ | (-1.024,1.024) | 2 | (1.024,1.024) |
| (0,0) | | (0,0) | 0 | |
| (2.048,0) | | (1.024,1.024) | 1 | |
| (2.048, 2.048) | | (1.024,1.024) | 0 | |

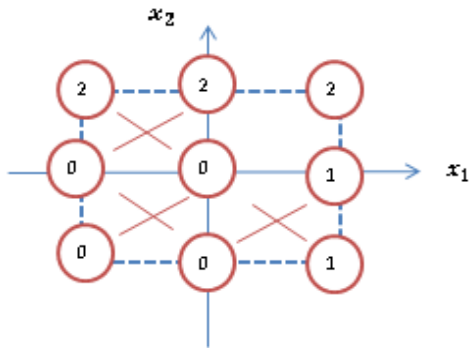

Table 14: The second generation of $f_2$

| $h_3 = 1.024$ P(2): | $h_4 = 0.512$ | P(3): | $l(x)$ | Solution |
|---|---|---|---|---|
| (1.024,2.048) | $\xrightarrow{M}$ | (1.536,2.048) | 0 | (1.024,1.024) |
| (2.048,1.024) | | (1.536,1.536) | 1 | |
| (1.024,1.024) | | (1.024,1.024) | 0 | |
| (1.024,0) | | (0.512,0) | 1 | |
| (0,1.024) | | (-0.512,0.512) | 2 | |

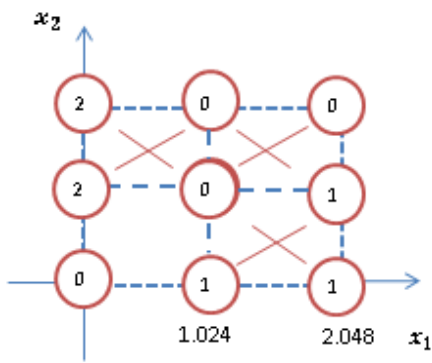

Figure 12: The second generation of $f_2$

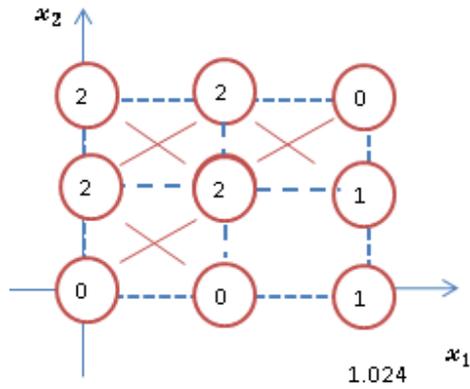

Figure 13: The Third generation of $f_2$

Table 15: The Third generation of $f_2$

| $h_4 = 0.512$ P(3): | $h_5 = 0.256$ | P(4): | $l(x)$ | Fixed point |
|---|---|---|---|---|
| (1.024,0.512) |  | (0.768,0.512) | **1** |  |
| (0.512,0) |  | (0.768,0.512) | **0** |  |
| (0,0.512) | $\xrightarrow{M}$ | (-0.256,0.256) | **2** | (1.024,1.024) |
| (0.512,0.512) |  | (0.512,0.256) | **2** |  |
| (0.512,1.024) |  | (0.512,0.768) | **2** |  |

Table 16: comparison between test problem #4 and other three methods

| **Algorithms** | **Iteration** | **Optimal point** | **Best point** | **Standard deviation** |
|---|---|---|---|---|
| **SLM** | 9 | (1.008,1.008) |  | (**0.008, 0.008**) |
| **RS** | 1000 | (1.062,1.062) |  | (0.062,0.062) |
| **RSW**($x^{initial}$ = (10.035,10.035) ) | 500 | (1.002,1.002) | (1,1) | (0.002,0.002) |
| **SA** | 150 | (1.0012,1.0012) |  | (0.0012, 0.0012) |

SLM of test problem 4 is compared with three other methods and their results are shown in table 16. Results show that the SLM earns the global optimal the same as other methods; but with lower steps and more precision than the other methods.

### 3.2.5 Test problem 5

This problem is Fifth De Jong function (Shekel's Foxholes).[6]

$$f_5(x_0, x_1) = \frac{1}{0.002+\sum_{i=0}^{24}\frac{1}{i+\sum_{j=0}^{1}(x_j-a_{ij})^6}}; \quad x_j \in [-65.536, 65.536] \tag{13}$$

With $a_{i0}=\{-32,-16,0,16,32\}$ for $i = 0,1,2,3,4$ and $a_{i0} = a_{i \bmod 5, 0}$

As well as $a_{i1}=\{-32,-16,0,16,32\}$ for $i = 0,5,10,15,20$ and $a_{i1} = a_{i+k,1}$,

$k = 1,2,3,4$. The global minimum for this function is $f_5(-32,-32) \cong 0.998004$.

Table 17: The initial population $f_5$

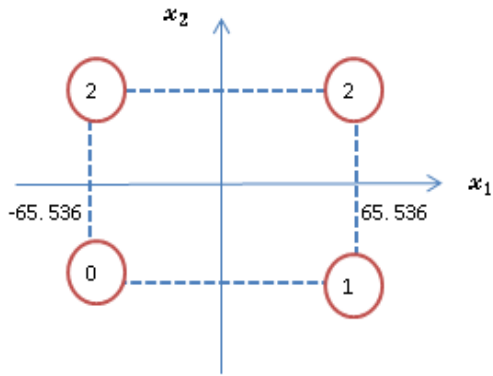

| $h_1 = 131.072$ P(0): | $h_2 = 65.536$ | P(1) : | $l(x)$ | Solution |
|---|---|---|---|---|
| (65.536, 65.536) | | (0,0) | 2 | |
| (65.536, -65.536) | $\xrightarrow{M}$ | (0,0) | 1 | (0,0) |
| (-65.536, 65.536) | | (0,0) | 2 | |
| (-65.536, -65.536) | | (0,0) | 0 | |

Figure 14: Initial Population of $f_5$

Table 18: The first generation of $f_5$

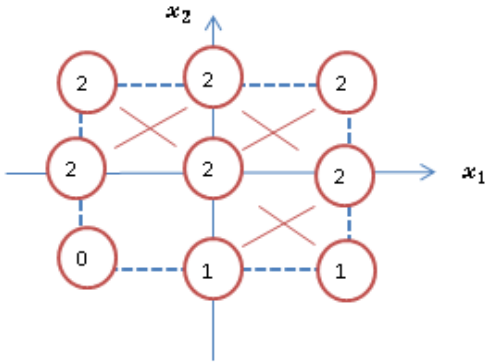

| $h_2 = 65.536$ P(1): | $h_3 = 32.768$ | P(2): | $l(x)$ | Solution |
|---|---|---|---|---|
| (65.536, 0) | | (32.768, -32.768) | 2 | |
| (0, 65.536) | | (-32.768, 32.768) | 2 | |
| (0, 0) | $\xrightarrow{M}$ | (-32.768, -32.768) | 2 | (-32.768, -32.768) |
| (-65.536, 0) | | (-32.768, -32.768) | 2 | |
| (0, -65.536) | | (-32.768, -32.768) | 1 | |

Figure 15: the first generation of $f_5$

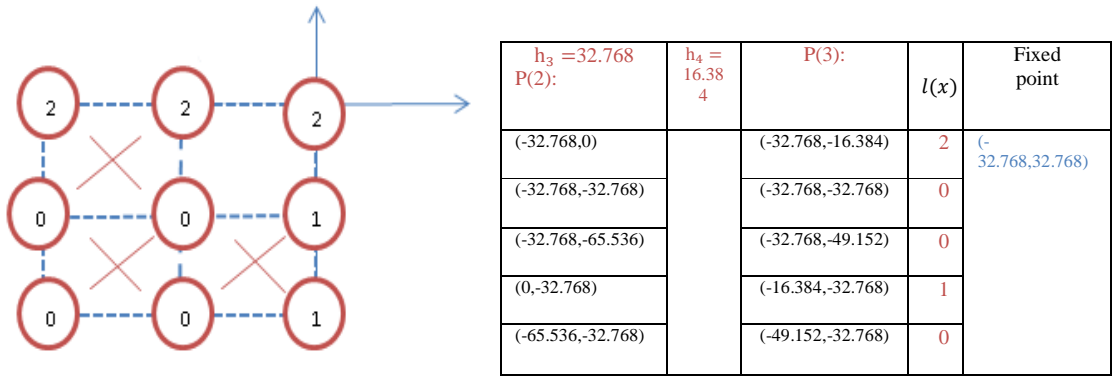

Table 19: The second generation of $f_5$

| $h_3 = 32.768$ P(2): | $h_4 = 16.384$ | P(3): | $l(x)$ | Fixed point |
|---|---|---|---|---|
| (-32.768,0) | | (-32.768,-16.384) | 2 | (-32.768,32.768) |
| (-32.768,-32.768) | | (-32.768,-32.768) | 0 | |
| (-32.768,-65.536) | | (-32.768,-49.152) | 0 | |
| (0,-32.768) | | (-16.384,-32.768) | 1 | |
| (-65.536,-32.768) | | (-49.152,-32.768) | 0 | |

Figure 16: The second generation of $f_5$

Table 20: comparison between test problem #5 and other three methods

| Algorithms | Iteration | Optimal point | Best point | Standard deviation |
|---|---|---|---|---|
| **SLM** | 9 | (-32.256, -32.256) | | (0.256, 0.256) |
| **RS** | 1000 | (-32.657, 32.657) | $(-32, -32)$ | (0.657, 0.657) |
| **RSW**($x^{initial} = (10.835, 20.835)$) | 500 | (-32.013, 32.012) | | (0.013, 0.012) |
| **SA** | 150 | (-32.0053, -32.0053) | | (0.0053, 0.0053) |

We compared the SLM of the test problem 5 with other methods that their results are shown in table 20. Results show that the SLM performs better than other methods point of view number of steps and precision.

## 4 Time Complexity

In this section, we calculate time complexity of the SLM and get T(n)=$14\log_2^n$ according to the following computetional .This Time complexity is equal to $14\log_2^n$ for all fucntions with different dimensional which is advantage compared with other methods

.In this computational, n is the number of dimensions and because space is divided 2 in each step then we have T (n/2). As for every point in search space, we have to calculate some calculation so 28 must be added to T(n) and we have T (n) =T (n/2) +28 .

**T (n) =T (n/2) +28** $\xrightarrow{n=2^K}$ T (n=$2^K$) = T(n=$\frac{2^K}{2}$) + 28 $\rightarrow$ T(n=$2^K$)-T(n=$2^K-1$) =28

$\xrightarrow{T(n=2^K)=T_K}$ $T_K - T_{K-1} = 28$ $\xrightarrow{T_K = X^P}$ $X^P - X^{P-1} = 0$ $\xrightarrow{P=1}$

X-1=0 $\longrightarrow$ $X_{1=1}$

$28=11^K * 28K^0$ $\xrightarrow[p(K)=K\ D=0]{\substack{P-1=0 \\ K=1\ ,}}$ $(X-1)^{0+1}=0$ $\longrightarrow$ $X_2 = 1$

$T_K = C_1 1^K + C_2 K 1^K$ $\xrightarrow{K=\log_2^n}$ $T(2^K) = C_1 1^K + C_2 K 1^K$ $\xrightarrow{N=2^k}$ $T(n) = C_1 + C_2 \log_2^n$

$C_1 + 2(14) = 28$ $\longrightarrow$ $t_4 = 28$

$\left\{ \vphantom{\begin{array}{c}1\\1\\1\\1\\1\\1\end{array}} \right. \Rightarrow \begin{array}{l} 2c_2 = 28 \\ c_2 = 14 \\ T(n) = 14\log_2^n \end{array}$

## 5 Conclusions:

In this paper, the SLM is proposed for the optimization problems and implemented on five optimization problems such as the Second and Fifth of De Jong functions.According to tables 1 to 5, the numerical results show that the SLM in multi-dimensional and great search space perform better than other previously proposed techniques (RS, RSW, SA) in terms of the number of steps and precision for finding global point .Also, its time complexity is optimal so proposed algorithm is efficient and reliable.